\documentclass[10pt, a4paper]{article}

\usepackage{lrec-coling2024} 

\usepackage{multirow}
\usepackage{enumitem}
\usepackage{makecell}
\usepackage{ragged2e}
\usepackage{booktabs} 
\usepackage{amsmath,amsthm,amssymb,amsfonts} 

\title{BP4ER: Bootstrap Prompting for Explicit Reasoning \\in Medical Dialogue Generation}

\name{Yuhong He\textsuperscript{1,6}, Yongqi Zhang\textsuperscript{2}, Shizhu He\textsuperscript{3,4} and Jun Wan\textsuperscript{1,3,5,*}\thanks{* Corresponding author}} 

\address{
\textsuperscript{1} Macau University of Science and Technology, Macao, China\\ 
\textsuperscript{2} 4Paradigm Inc., Beijing, China \\ \textsuperscript{3} School of Artificial Intelligence, University of Chinese Academy of Sciences, Beijing, China\\
\textsuperscript{4} The Lab of Cognition and Decision Intelligence for Complex Systems, CASIA, China \\
\textsuperscript{5} MAIS, Institute of Automation, Chinese Academy of Sciences, Beijing, China\\
\textsuperscript{6} Zhongkai University of Agriculture and Engineering, Guangzhou, China\\
yuhonghe.ai@gmail.com, 
yzhangee@connect.ust.hk, shizhu.he@nlpr.ia.ac.cn, 
jun.wan@ia.ac.cn
}

\abstract{
Medical dialogue generation (MDG) has gained increasing attention due to its substantial practical value. Previous works typically employ a sequence-to-sequence framework to generate medical responses by modeling dialogue context as sequential text with annotated medical entities. While these methods have been successful in generating fluent responses, they fail to provide process explanations of reasoning and require extensive entity annotation. To address these limitations, we propose the method \textbf{B}ootstrap \textbf{P}rompting for \textbf{E}xplicit \textbf{R}easoning in MDG (\textbf{BP4ER}), which explicitly model MDG’s multi-step reasoning process and iteratively enhance this reasoning process. We employ a least-to-most prompting strategy to guide a large language model (LLM) in explicit reasoning, breaking down MDG into simpler sub-questions. These sub-questions build on answers from previous ones. Additionally, we also introduce two distinct bootstrapping techniques for prompting, which autonomously correct errors and facilitate the LLM’s explicit reasoning. This approach eliminates the need for entity annotation and increases the transparency of the MDG process by explicitly generating the intermediate reasoning chain. The experimental findings on the two public datasets indicate that BP4ER outperforms state-of-the-art methods in terms of both objective and subjective evaluation metrics.
 \\ \newline  }

\begin{document}

\maketitleabstract

\section{Introduction}
\label{intro}
Medical dialogue systems (MDS) are receiving significant attention due to the rising demand for telemedicine \citep{zhou2021generation,he2022Dialmed}, offering accessible medical services such as health consultations, diagnosis, and prescriptions, to a broader population \citep{yan2022remedi,xia2022medconqa}. Within MDS, medical dialogue generation (MDG) plays a crucial role by generating accurate medical responses based on given dialogue histories \cite{lin2021aaai,wei2018acl,xu2019aaai}. Typically, MDG involves understanding the patient's overall state, making the next diagnosis decisions in a limited-turn dialogue, and conducting medical reasoning analysis to generate responses \cite{li2021sigir,chen2022aaai}.

\begin{figure}
\begin{center}
\includegraphics[width=0.49\textwidth]{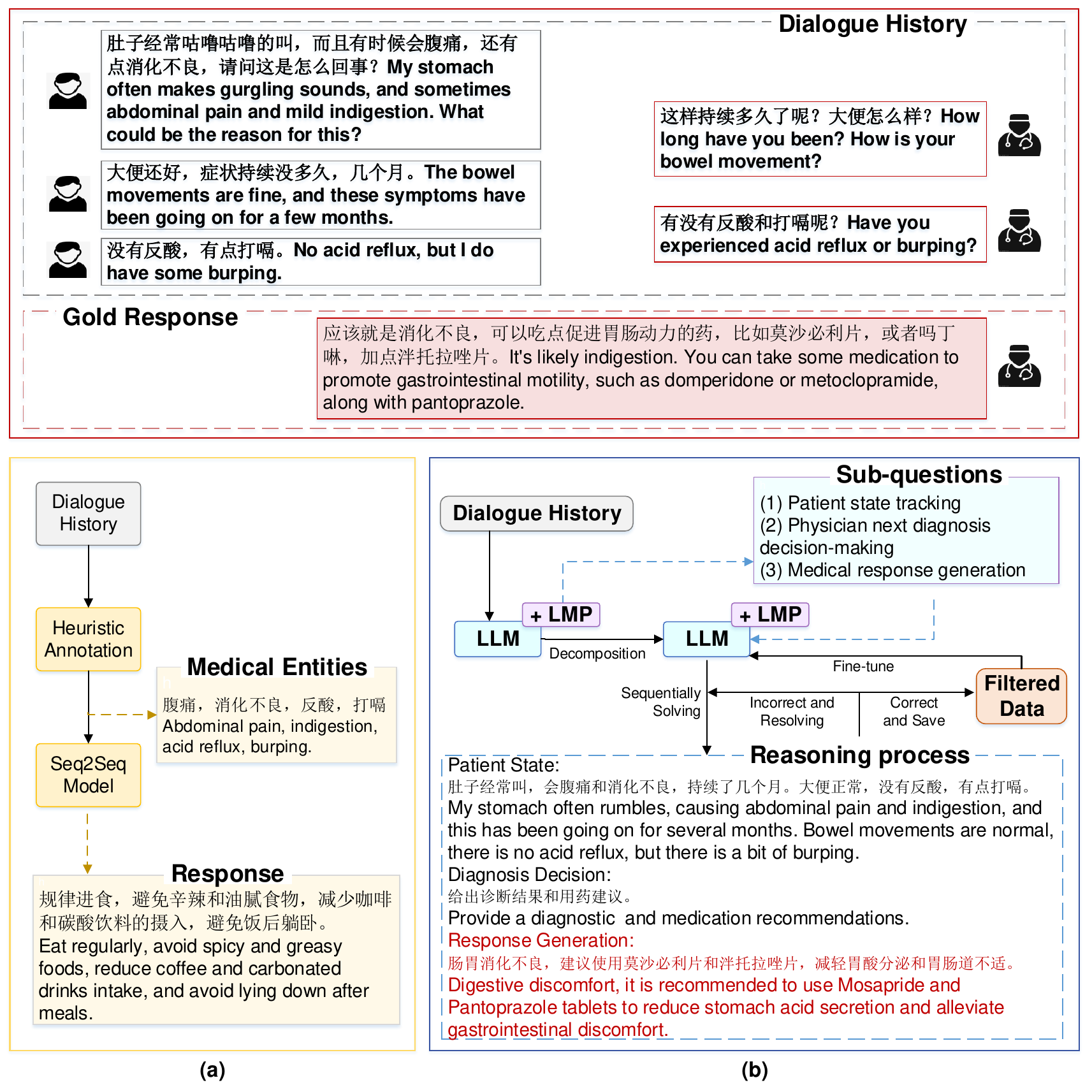} 
\caption{Paradigm comparison in MDG: prior works adopt a Seq2Seq framework (a); our model (b) explicitly incorporates a multi-step reasoning process and reduces entity annotation.}
\label{fig_intro}
\end{center}
\end{figure}

Previous research on MDG typically adopts a framework in which dialogue context is modeled as sequential text \cite{xu2023acl,liu2021nc}, and medical entities are identified and annotated within this textual context \cite{liu2020arxiv,du2019emnlp}. Subsequently, response generation is carried out using sequence-to-sequence (Seq2Seq) models \cite{sutskever2014sequence}. These Seq2Seq methods leverage pre-trained text encoders and decoders to generate medical responses \cite{li2021sigir,zhao2022kdd}, as illustrated in Figure \ref{fig_intro} (a). Although these methods have yielded substantial success in generating coherent and fluent responses in MDG, they face two key challenges: (1) \emph{Lack of process explanation.} To help patients or physicians understand why an MDG module generates a response, interpretability of the medical reasoning process is indispensable \cite{li2021sigir}, i.e., information on patient status and diagnostic decision-making by physicians. (2) \emph{Requirement for large-scale annotations.} Previous works \cite{xu2023acl,zhao2022kdd} heavily depend on the availability of a substantial amount of manually labeled data during the training phase. However, obtaining such data is often challenging due to the specialized medical knowledge required and stringent privacy considerations.

To address the limitations above, we propose the \textbf{B}ootstrap \textbf{P}rompting for \textbf{E}xplicit \textbf{R}easoning method (\textbf{BP4ER}), as illustrated in Figure \ref{fig_intro} (b). Our motivation is to eliminate the need for entity annotation by treating MDG as a multi-step reasoning problem. Specifically, we explicitly break down MDG into a reasoning chain and sequentially address each intermediate reasoning step, aligning with its inherent multi-step reasoning process. Drawing from the concept of chain-of-thought prompting \cite{wei2022chain}, we introduce the least-to-most prompting (LMP) strategy \cite{zhou2023iclr} to guide a large language model (LLM) \cite{zhao2023survey,du2022glm} towards explicit reasoning in MDG. We first decompose the MDG process into a reasoning chain, comprising a series of interrelated sub-questions. Then, we follow \citet{zelikman2022nips} and construct demonstration prompts for each sub-question and address them sequentially with answers from resolved sub-questions, promoting a coherent reasoning process.

Despite LLMs' impressive language understanding ability in general language modeling \cite{wang2022language,huang2022towards,zhu2023learning}, their intermediate reasoning steps in MDG would be error-prone, reducing overall performance \cite{zhang2022automatic}. To facilitate the model's explicit reasoning ability, we propose two distinct bootstrapping techniques for prompting: answer-providing bootstrapping (AP-Bootstrap) and prompt-revising bootstrapping (PR-Bootstrap). These techniques allow the model to autonomously rectify errors without relying on large-scale annotations. Subsequently, we collect the accurate reasoning chain to create filtered data by implementing feedback loops. The model is then fine-tuned using this filtered data, and the process is repeated. This approach yields a significant improvement in the model's performance and enhances the quality of the generated responses. 

Our contributions can be summarized as follows:

\begin{itemize}
\item We present a novel explicit reasoning model for medical dialogue generation (MDG) called BP4ER. To the best of our knowledge, BP4ER is the first model to systematically deconstruct MDG into an intermediate reasoning chain, which notably enhances the interpretability of the MDG process.
\item BP4ER introduces the least-to-most prompting strategy to guide LLM for explicit reasoning and an iterative approach to bootstrap the prompting process for augmenting the LLM's reasoning capabilities, resulting in coherent and precise medical dialogue responses.
\item  We evaluate BP4ER on two public datasets using both automatic and manual evaluation metrics. Experimental results demonstrate its superiority over previous methods.
\end{itemize}

\begin{figure*}
\begin{center}
\includegraphics[width=\textwidth]{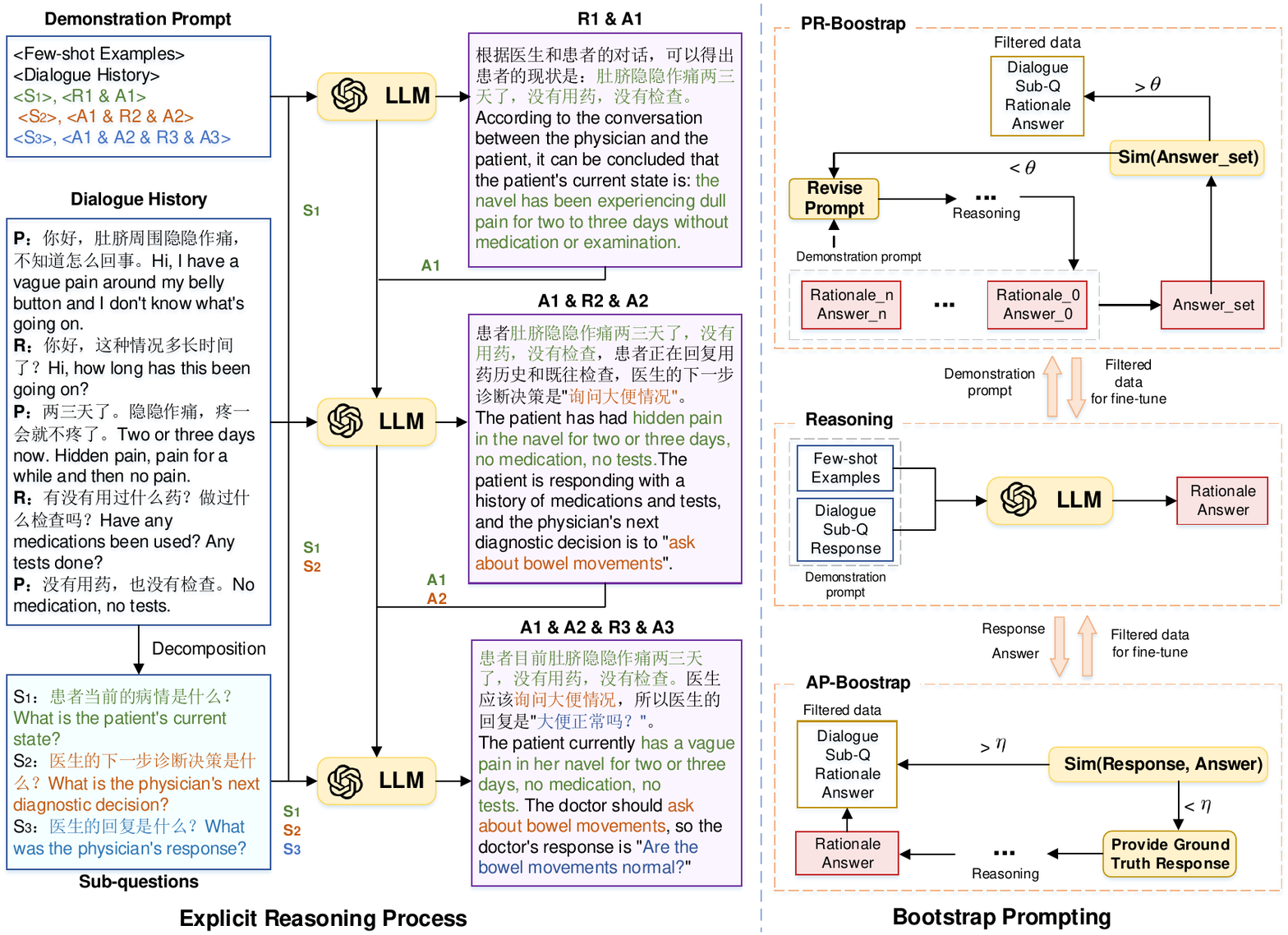} 
\caption{Overview of BP4ER. Medical dialogue is deconstructed into a reasoning chain of sub-questions. Demonstration prompts guide intermediate reasoning, sequentially querying the LLM. Two bootstrapping techniques for prompting, AP-Bootstrap and RP-Bootstrap, are introduced to enhance explicit reasoning.}
\label{fig_main}
\end{center}
\end{figure*}

\section{Related Work}
\subsection{Medical Dialogue Generation}
Medical dialogue generation (MDG) has attracted increasing attention due to its high practical value. Early attempts at MDG were based on pre-defined templates to generate natural language \cite{Ferguson2009CARDIACAI,wong2011health,xu2019end}. However, template-based MDG suffers from the problem of inflexibility. Recently, \citet{zeng2020emnlp} took an initial step in neural-based MDG. They pre-trained several dialogue generation models on large-scale medical corpora. \citet{liu2020arxiv} frame medical dialogue generation as entity prediction and entity-aware response generation. Furthermore, \citet{liu2021nc} unifies the dialogue context understanding and entity reasoning through a heterogeneous graph. \citet{li2021sigir} consider medical entities in the utterances as states and actions and present semi-supervised variation reasoning with a patient state tracker and a physician action network. \citet{zhao2022kdd} exploit the medical relationship between dialogue context and recall pivotal information to produce responses. \citet{xu2023acl} models a medical entity flow and a dialogue act flow to improve entity selection and dialogue act prediction. 

Although these models achieve comparable performance, they often lack process interpretability and need substantial annotation.
\subsection{Prompt Learning of LLMs}
Recent studies \cite{dong2023survey,jeblick2022chatgpt} have proposed various prompting strategies to strengthen and generalize the in-context learning ability of LLMs. One such strategy is chain-of-thoughts (CoT) prompting, introduced by \citet{wei2022chain}, which incorporates intermediate reasoning steps into LLMs to construct demonstrations between inputs and outputs. While \citet{wei2022chain} manually constructs CoTs, AutoCoT \cite{zhang2022automatic} utilizes LLMs to automatically generate CoTs, using the prompt sentence "let’s think step by step." Additionally, \citet{wang2022iteratively} propose iCAP, a context-aware prompter capable of dynamically adjusting contexts for each reasoning step. To tackle the challenge of easy-to-hard generalization, \citet{zhou2023iclr} propose a least-to-most prompting (LMP) strategy. Unlike CoT, which focuses on individual instances, LMP is task-oriented, breaking down a problem into interrelated sub-questions from a task perspective and forming a progressive prompt sequence for LLMs. Moreover, while CoTs are crucial for model performance, they are not readily available for specific tasks, and creating them requires significant time and resources, potentially introducing bias. 

Inspired by LMP, we introduce MDG as a multi-step reasoning problem aimed at explicitly and iteratively modeling the reasoning process, mirroring the decision-making process of doctors in real medical scenarios.

\section{Main Method}
\paragraph{Problem Formulation.}
In the context of a dialogue comprising $T$ turns, a medical dialogue session $D$ is a sequence of utterances, denoted as $D = \{P_1, R_1, P_2, R_2,..., P_T, R_T\}$. Here, $P_t$ and $R_t$ ($t=1\dots T$) refer to utterances from a patient and responses from a virtual physician, respectively. At the $t$-th turn, given the dialogue history $H = \{P_1, R_1,...R_{t-1}, P_t\}$ as input, the model aims to generate an intermediate reasoning chain $S=\{S_1, ..., S_k\}$ and corresponding answers $A = \{A_1, ..., A_k\}$, where $k$ is the number of reasoning steps. Subsequently, the model generates a medical response $R_t$ for the current turn. Figure \ref{fig_main} provides an illustrative overview of our proposed BP4ER method. In this section, we provide a description of the multi-step reasoning process for MDG, as outlined in Section \ref{MR}. Then, we present the details of the explicit reasoning process in Section \ref{ER}, with a specific focus on augmenting the model's interpretability. Finally, we introduce two distinct bootstrapping techniques for prompting to enhance explicit reasoning in the BP4ER model, as discussed in Section \ref{BP}.

\subsection{Multi-step Reasoning}
\label{MR}
In real-world medical scenarios, MDG involves a multi-step reasoning process that aligns with the logical framework of medical consultation \cite{chen2022aaai}. It consists of three essential steps \cite{li2021sigir}:
(i) \emph{Patient State Tracking:} Initially, the MDG system interacts with the patient to acquire additional symptoms beyond those self-reported. Here, the system focuses on comprehensively tracking and maintaining the patient's condition within the dialogue context, including symptoms, medications, and other relevant information.
(ii) \emph{Next Diagnosis Decision-making:} Drawing from the collected patient states and the ongoing conversation, the system infers the next diagnosis decision that a physician would make. This step guides the responses generated by the system, ensuring a coherent flow in the medical dialogue.
(iii) \emph{Medical Response Generation:} Utilizing the identified patient states and the diagnosis decision-making, the MDG system generates a contextually relevant and coherent response that aligns with the ongoing medical dialogue.

\subsection{Explicit Reasoning Process}
\label{ER}
In Section \ref{intro}, we emphasized the importance of explicitly demonstrating the multi-step reasoning process of MDG for better interpretability, rather than simply generating direct answers. 
To achieve this, we employ a few-shot Least-to-Most Prompting (LMP) strategy \cite{zhou2023iclr} to guide the Large Language Model (LLM) \cite{zhao2023survey,du2022glm}. This strategy breaks down the complex MDG task into a sequence of interrelated sub-questions, inspired by medical diagnostic logic. In this study, we simplify this decomposition into three specific sub-questions following the multi-step reasoning process described in Section \ref{MR}, creating an intermediate reasoning chain, denoted as $S=\{S_1, S_2, S_3\}$:

\begin{itemize}
    \item $S_1$: What's the patient's current state?
    \item $S_2$: What's the physician's next decision?
    \item $S_3$: What's the physician's response? 
\end{itemize}
As depicted in Figure \ref{fig_main}, the process of generating a response from a dialogue history is reframed as answering two intermediate sub-questions: "\emph{What's the patient's current state?}" and "\emph{What's the physician's next diagnostic decision?}". 

We tackle these sub-questions sequentially, with each solution building upon previously obtained answers. To facilitate this, we create question-rational-answer pairs as demonstrations and construct a demonstration prompt for each intermediate reasoning step, inspired by \cite{zelikman2022nips}. This prompt consists of examples illustrating sub-question resolution, the dialogue history, a list of previously answered sub-questions and their corresponding answers (if any), and the next sub-question to be addressed.

The solving process starts with a few-shot prompting, providing the LLM with a demonstration prompt comprising few-shot examples, dialogue history, and the first sub-question. For example in Figure \ref{fig_main}, the demonstration prompt is "\emph{Examples: <Few-shot Examples>, 
Dialogue History $H$: P: Hi, I have a vague pain ... P: No medication, no tests, 
Sub-question $S_1$: What's the patient's current state?}". Then, We use the generated answer, e.g., "$A_1$: \emph{The navel ... examination}," to construct the next prompt by appending the answer to the previous prompt followed by the next sub-question $S_2$: "\emph{What's the physician's next diagnostic decision?}". This process repeats for sub-question $S_3$: "\emph{What's the physician's response?}". The final answer (e.g., "$A_3$: \emph{Are the bowel movements normal?}") for MDG $R_t$ is obtained by adding the generated answer $A_2$ to the previous prompt. This approach allows us to address each sub-question sequentially, leveraging answers from previously resolved sub-questions, resulting in a coherent, step-by-step reasoning process.

\subsection{Bootstrap Prompting}
\label{BP}
The intermediate reasoning steps in LLMs may contain errors, affecting reasoning results and overall performance. To enhance the explicit reasoning abilities of LLMs, drawing inspiration from \cite{wang2022iteratively}, we improve the quality of demonstrations through iterative prompting bootstrapping. During the training phase, the final step of reasoning benefits from having access to ground truth responses, ensuring accuracy. However, intermediate steps lack correct answers, posing a challenge. To overcome this limitation, we introduce two iterative bootstrapping techniques for prompting: answer-providing bootstrapping (AP-Bootstrap) and prompt-revising bootstrapping (PR-Bootstrap), tailored to different scenarios. AP-Bootstrap can be seen as a greedy decoding process, whereas PR-Bootstrap is based on a sampling approach. These techniques help LLMs to autonomously rectify errors in demonstrations, reducing the reliance on extensive annotations.


\subsubsection{Answer-Providing Bootstrapping}
Given a pre-trained LLM $\mathcal{M}$ and a dataset of dialogue histories $\mathcal{H}$ paired with responses $\mathcal{R}$, denoted as $\mathcal{D}=\{(H_i, R_i)\}_{i=1}^{N_{D}}$, the AP-Bootstrap approach takes a demonstration prompt as input. This prompt consists of a small example set $\mathcal{P}$, defined as $\mathcal{P}=\{H^p_i,Q^p_i,R^p_i\}_{i=1}^{N_{P}}$, where $N_{P} \ll N_{D}$ (e.g. $N_{P}=5$). Similar to standard few-shot prompting, this example set is concatenated with each dialogue history instance in $\mathcal{D}$ and sub-question $S_i$, resulting in $\hat{H}_i=\{H^p_1, Q^p_1, R^p_1,..., H^p_{N_{P}}, Q^p_{N_{P}}, R^p_{N_{P}}, H_i, S_i \}$. This encourages the model to generate a rationale $\hat{Q}_i$ for $H_i$ followed by an answer $A_i$. If the generated answers $A_i$ are semantically similar to the gold response $R_i$, the reasoning process is considered credible. Otherwise, our objective is to correct the reasoning process and obtain available answers $A_i$. Finally, the credible and corrected dialogue data are combined for iterative fine-tuning of the LLM, enhancing its reasoning capabilities.
 
To achieve this, we employ cosine similarity, denoted as $Sim(.)$, to measure the semantic similarity between the generated answers $A_i$ and the gold response $R_i$. We utilize this similarity metric to filter the dialogue data, retaining instances with high semantic similarity, i.e., $Sim(A_i, R_i) > \eta$, where $\eta$ is a predefined threshold. For those instances with low similarity, following \cite{zelikman2022nips}, we provide a model with the gold response, allowing it to autonomously rectify errors by generating a reasoning chain similar to the previous explicit reasoning process (as described in Section \ref{ER}). By providing the gold response, the model can reason backward, facilitating the generation of a reasoning chain leading to the correct answer. After error correction, the dialogue with the revised reasoning chain is added to the filtered dataset. Subsequently, we fine-tune the LLM $\mathcal{M}$ on this filtered dataset and iteratively bootstrap prompting $\mathcal{M}$ to generate a new reasoning chain with the newly fine-tuned model until performance reaches a plateau. Throughout this iterative process, we consistently fine-tune from the original pre-trained model $\mathcal{M}$ to mitigate overfitting concerns.

The AP-Bootstrap method can be conceptualized as an approximation to an RL-style policy gradient objective. To illustrate this, consider that $\mathcal{M}$ can be interpreted as a discrete latent variable model $p_M(R|H)=\sum_Q p(Q|H)p(R|H, Q)$; in other words, $\mathcal{M}$ first samples a latent rationale $Q$ before generating the response $R$. Now, given the indicator reward function $f_I=\mathbb{I}(Sim(A,R)>\eta)$, the total expected reward across the dataset is:
\begin{equation*}
\mathcal{J}(\mathcal{M},H,R) = \sum\nolimits_i \mathbb{E}_{\hat{Q}_i,A_i\sim p_M(\cdot|H_i)}f_I(\cdot)
\end{equation*}
whose gradient is obtained via the standard log-derivative trick for policy gradients:

\begin{multline*}
    \nabla \mathcal{J}(\mathcal{M},H,R) = \sum\nolimits_i \mathbb{E}_{\hat{Q}_i,A_i \sim p_M(\cdot|H_i)} \\ 
    [f_I(\cdot) \cdot \nabla log p_M=(A_i,\hat{Q}_i|H_i)]
\end{multline*}
Note that the indicator function discards the gradient for dissimilar sampled demonstrations to the correct response $R_i$. Thus, the AP-Bootstrap approximates $\mathcal{J}$ by 1) greedily decoding samples of $(\hat{Q}_i, A_i)$ to reduce the variance of this estimate, and 2) taking multiple gradient steps on the same data batch, similar to policy gradient algorithms \cite{schulman2017proximal}.

\subsubsection{Prompt-Revising Bootstrapping}
During our experiments, we noticed that autonomous error correction faces challenges when dealing with complex dialogues, such as doctors continuously questioning patients, cross-questions between doctors and patients, and ambiguous descriptions of patient conditions. We attribute this challenge to the lack of correct answers in the intermediate steps of the reasoning process within MDG. To address this, we introduce a straightforward yet effective strategy called prompt-revising bootstrapping (PR-Bootstrap). This strategy capitalizes on the understanding that complex reasoning tasks often offer multiple pathways to arrive at a correct answer, as discussed in \cite{stanovich200024}. In contrast to AP-Bootstrap, PR-Bootstrap alleviates the problem of limited diversity inherent in greedy decoding, as demonstrated in our experiments.

To implement PR-Bootstrap, we first prompt the LLM in the format of a demonstration prompt to yield an initial answer, which is added to the candidate answers. We then revise the few-shot examples in the original demonstration prompt to generate an alternative rationale, along with its corresponding new answer, which is also included in the candidate answers. It's important to note that each answer within the candidate set is derived from a distinct rationale. Therefore, if two answers exhibit significant semantic similarity, they are considered a consistent answer pair. We measure this similarity using cosine similarity calculations between the newly generated answer and those in the candidate set. Answer pairs surpassing a predefined threshold $\theta$ are considered the most consistent within the candidate answer set and are added to the filtered dataset. When no answer pairs meet the threshold $\theta$, we iterate the prompt revision process to explore diverse reasoning paths and generate alternative answers until reliable answers are obtained for all provided data. 

The iterative bootstrapping approach mirrors the human experience, where multiple different reasoning paths leading to the same answer increase confidence in its correctness. Finally, similar to the AP-Bootstrap method, we fine-tune the LLM on the filtered dataset to enhance its reasoning abilities by bootstrapping the prompting process.

\section{Experiments}
\subsection{Datasets}
We adopt two publicly available benchmark datasets, namely MedDG \citelanguageresource{meddg} and KaMed \citelanguageresource{kamed}, collected from medical consultation websites\footnote{\url{https://www.chunyuyisheng.com/}} after anonymization.
\textbf{MedDG} contains 17K dialogues, focusing on 12 distinct diseases within the gastroenterology department. On average, each dialogue consists of 9.92 rounds. We divide the dataset into training/validation/test sets with sizes of 14,864/2,000/1,000 dialogues, as originally outlined in \citet{liu2020arxiv}.
\textbf{KaMed} contains over 63K dialogues, covering an extensive range of over 300 diseases across 13 different medical departments. KaMed exhibits a higher average dialogue length compared to MedDG, e.g., 11.62 rounds per dialogue. Following the setting in \citet{xu2023acl}, we filtered dialogues with privacy concerns and obtained 29,159/1,532/1,539 dialogues for the training/validation/test sets. The dataset presents challenging and diverse scenarios, with over 300 hospital departments.

\subsection{Evaluation metrics}
\textbf{Automatic Evaluation.}
To evaluate the linguistic quality of the generated responses, we employ standard word-overlap-based metrics: BLEU (B@n) \cite{papineni2002acl} and ROUGE (R@n) \cite{lin2004tsbo}. These metrics measure lexical quality by calculating n-gram overlaps between the generated and accurate responses. Additionally, we incorporate the DISTINCT (D@n) metric \cite{li2016naacl} for a more comprehensive evaluation. DISTINCT-n measures response diversity by calculating the proportion of distinct n-grams within the generated responses, offering a valuable perspective on response quality often missed by traditional BLEU and ROUGE metrics.

\noindent\textbf{Human Evaluation.} 
Aligned with prior studies \cite{li2021sigir, zhao2022kdd}, we conducted a human evaluation to assess the quality of responses in terms of fluency, coherence, and correctness. Fluency evaluation measures overall smoothness and naturalness, coherence assesses logical consistency with the dialogue history, and correctness measures the accuracy of medical knowledge in the responses. Consistent with \citet{li2021sigir} and \citet{zhao2022kdd}, we randomly sampled 100 cases and invited three professional annotators from a thirty-party hospital to perform manual evaluations. Annotators utilized the aforementioned metrics, rating each response on a scale from 1 (poor) to 5 (excellent). It's noteworthy that model names were anonymized to ensure objectivity throughout the evaluation process.

\begin{table*}[ht]
\centering
\begin{tabular}{llllllll}
\hline
\textbf{Dataset} & \textbf{Model} & \textbf{B@1} & \textbf{B@2} & \textbf{B@4} & \textbf{R@1} & \textbf{R@2} & \textbf{D@2} \\
\hline
\multirow{8}*{MedDG} & Seq2Seq \cite{sutskever2014sequence} & $28.55$ & $22.85$ & $15.45$ & $25.61$ & $11.24$ & / \\
& HRED \cite{serban2016building} & $31.61$ & $25.22$ & $17.05$ & $24.17$ & $9.79$ & / \\
& DialoGPT \cite{zhang2019dialogpt} & $32.77^\dagger$ & $26.93^\dagger$ & $17.96^\dagger$ & $27.11^\dagger$ & $11.34^\dagger$ & $79.26^\dagger$ \\
& GPT-2 \cite{radford2019language} & $35.27$ & $28.19$ & $19.16$ & $28.74$ & $13.61$ & / \\
& VRBot \cite{li2021sigir} & $29.69$ & $23.9$ & $16.34$ & $24.69$ & $11.23$ & / \\
& MedPIR \cite{zhao2022kdd} & $38.72^\dagger$ & $27.64^\dagger$ & $18.14^\dagger$ & $25.72^\dagger$ & $10.30^\dagger$ & $82.77^\dagger$ \\
& DFMed \cite{xu2023acl} & \underline{$42.56$} & \underline{$33.34$} & \underline{$22.53$} & \underline{$29.31$} & \underline{$14.21$} & / \\
& ChatGLM-6B \cite{du2022glm} & $37.96$ & $24.22$ & $15.37$ & $18.05$ & $10.53$ & \underline{$89.81$} \\  \cline{2-8}
& BP4ER (ours) & \textbf{44.78} & \textbf{33.80} & \textbf{23.76} & \textbf{41.47} & \textbf{22.47} & \textbf{89.93} \\ 
& Improvement & +2.22 & +0.46 & +1.23 & +12.16 & +8.26 & +0.12 \\

\midrule
\multirow{8}*{KaMed} & Seq2Seq \cite{sutskever2014sequence} & $23.52$ & $18.56$ & $12.13$ & $23.56$ & $8.67$ & / \\
& HRED \cite{serban2016building} & $26.75$ & $21.08$ & $16.36$ & $18.71$ & $7.28$ & / \\
& DialoGPT \cite{zhang2019dialogpt} & $30.17^\dagger$ & $25.53^\dagger$ & $17.09^\dagger$ & $24.30^\dagger$ & $9.79^\dagger$ & $80.27^\dagger$\\
& GPT-2 \cite{radford2019language} & $33.76$ & $26.58$ & $17.82$ & $26.8$ & $10.59$ &/ \\
& VRBot \cite{li2021sigir} & $30.04$ & $23.76$ & $16.36$ & $18.71$ & $7.28$ & / \\
& MedPIR \cite{zhao2022kdd} & $29.42^\dagger$ & $21.60^\dagger$ & $16.47^\dagger$ & $20.69^\dagger$ & $9.27^\dagger$ & $83.75^\dagger$ \\
& DFMed \cite{xu2023acl} & \underline{$40.20$} & \underline{$30.97$} & \underline{$20.76$} & $28.28$ & $11.54$ & /\\
& ChatGLM-6B \cite{du2022glm} & $38.70$ & $27.19$ & $16.38$ & \underline{$33.86$} & \underline{$20.21$} & \underline{$85.70$} \\  \cline{2-8}
& BP4ER (ours) & \textbf{41.89} & \textbf{31.74} & \textbf{20.81} & \textbf{35.76} & \textbf{21.19} & \textbf{86.83}\\ 
& Improvement & +1.69 & +0.77 & +0.05 & +1.90 & +0.98 & +1.07 \\
\hline
\end{tabular}
\caption{\label{main_exam} Automatic evaluation (\%) on MedDG and KaMed datasets. B@n denotes BLEU-n, R@n denotes ROUGE-n and D@2 denotes DISTINCT-2. The best values are in boldface and the second best are underlined. Models marked with $\dagger$ were reproduced by us, while the others were copied from the original results in \cite{xu2023acl}.}
\end{table*}

\subsection{Implementation Details}
In this work, we used ChatGLM-6B\footnote{It can be done with any off-the-shelf LLMs, such as LLaMA \cite{touvron2023llama} and Alpaca \cite{taori2023alpaca}. } \cite{du2022glm} as the foundational LLM for BP4ER. ChatGLM-6B is equipped with 6 billion parameters and is optimized with the Adam optimizer \cite{kingma2014adam}. We chose this model for its robust language understanding abilities in Chinese and its relatively lightweight design compared to other LLMs. Hyperparameters were selected based on the best-performing checkpoints during validation, with a batch size of 32 and a learning rate of 1e-2. For MedDG, we set similarity thresholds as [0.75, 0.8, 0.65] for its three reasoning steps, while for KaMed, they were [0.65, 0.75, 0.65]. All experiments were conducted on a single NVIDIA GeForce RTX 3090 GPU.

\subsection{Baseline models}
Our method is compared with the following baselines. \textbf{Seq2Seq} \cite{sutskever2014sequence} is an RNN-based sequence-to-sequence model with an attention mechanism. \textbf{HRED} \cite{serban2016building} uses hierarchical encoders to model the dialogue context from token level and utterance level compared to Seq2Seq. \textbf{DialoGPT} \cite{zhang2019dialogpt} and \textbf{GPT-2} \cite{radford2019language} are transformers-based pre-trained language models widely adopted in tasks of dialogue generation. \textbf{VRBot} \cite{li2021sigir} summarizes patient states and physician actions into phrases through variational methods and generate the response. \textbf{MedPIR} \cite{zhao2022kdd} exploit the medical relationship between dialogue context and recall pivotal information to produce responses in the recall-enhanced generator. \textbf{DFMed} \cite{xu2023acl} models the transitions of medical entities and dialogue acts with the pre-trained model. \textbf{ChatGLM-6B} \cite{du2022glm} is a pre-trained language model with 6 billion parameters, which generates medical responses directly.

\subsection{Automatic Evaluation}
Table \ref{main_exam} presents the automatic evaluation results for the MedDG and KaMed datasets, revealing several key insights:

(1) BP4ER demonstrates significant improvements, as depicted in Table \ref{main_exam}. These results confirm BP4ER's efficacy in enhancing response quality and ensuring greater semantic consistency with gold standard responses. While ChatGLM-6B initially exhibits lower BLEU and ROUGE scores compared to DFMed, integrating explicit reasoning and bootstrapping prompting techniques yields notable enhancements. Specifically, there's a remarkable increase of $23.42\%$ in ROUGE-1 and $11.94\%$ in ROUGE-2. This integration not only boosts performance metrics but also enhances the transparency of the multi-step reasoning process in MDG. It renders the reasoning steps more comprehensible and interpretable without the need for extensive annotations. As a result, the model's decision-making process becomes more transparent, facilitating a deeper understanding of the underlying logic behind the generated responses.

(2) The performance enhancement is more pronounced in MedDG compared to KaMed, as indicated in Table \ref{main_exam}. This discrepancy can be attributed to the fact that MedDG is focused on a specific department, i.e., gastroenterology, and contains a relatively small number of diseases, only 12 diseases. In contrast, KaMed covers a more extensive range of over 300 diseases across 13 different medical departments. The diversity and complexity inherent in KaMed render it a more challenging dataset for BP4ER. Additionally, it's worth noting that KaMed involves a greater number of dialogue rounds compared to MedDG, suggesting that the necessity to consider larger contextual information contributes to the dialogue's complexity. In summary, this observation suggests that BP4ER demonstrates more effectiveness when confronted with smaller and more focused datasets like MedDG, and it may encounter greater challenges when confronted with larger and more diverse datasets featuring extended dialogue rounds, such as KaMed.

\begin{table*}[ht]
\centering
\begin{tabular}{cccc|cccc|cccc}
\hline
\textbf{Fine-} & \textbf{Exp.} & \textbf{AP-} & \textbf{PR-} & \multicolumn{4}{c}{\textbf{MedDG}} & \multicolumn{4}{|c}{\textbf{KaMed}} \\
\textbf{Tune} & \textbf{Rea.} & \textbf{Boots.} & \textbf{Boots.} & \textbf{B@1} & \textbf{R@1} & \textbf{D@1} & \textbf{D@2} & \textbf{B@1} & \textbf{R@1} & \textbf{D@1} & \textbf{D@2}\\
\hline
$\checkmark$ & $\checkmark$ & $\checkmark$ & $\checkmark$ & 44.78 & 41.47 & 91.20 & 89.93 & 41.89 & 35.76 & 89.10 & 86.83 \\
$\checkmark$ & $\checkmark$ & $\checkmark$ & & 42.27 & 37.64 & 89.76 & 88.73 & 40.69 & 35.01 & 87.97 & 85.94 \\
$\checkmark$ & $\checkmark$ && & 40.75 & 36.63 & 90.14 & 88.90 & 39.68 & 34.99 & 88.47 & 86.07 \\
$\checkmark$ &&& & 39.41 & 27.38 & 88.54 & 89.81 & 39.13 & 33.97 & 87.34 & 85.83 \\
\hline
\end{tabular}
\caption{\label{ablation} Ablation studies (\%) are carried out on two datasets by individually removing modules PR-Bootstrap, AP-Bootstrap and explicit reasoning process.}
\end{table*}

\begin{table}[ht]
\centering
\begin{tabular}{lccc}
\hline
\textbf{Model} & \textbf{Fluency} & \textbf{Cohe.} & \textbf{Correct.} \\
\hline
DialoGPT & 3.11 & 2.56 & 2.89 \\
MedPIR & 3.34 & 3.07 & 3.23 \\
BP4ER & 4.00 & 3.50 & 3.52 \\
\hline
Gold & 4.32 & 4.17 & 4.41 \\
\hline
\end{tabular}
\caption{Human evaluation (\%) results on KaMed. The maximum score for each indicator is 5.}
\label{human_eval}
\end{table}

(3) In comparison to traditional seq2seq-based models, LLM-based models demonstrate superior performance in the ROUGE and DISTINCT-2 metrics, while seq2seq-based models perform well on the BLEU metric. For instance, BP4ER achieves substantial improvements of $12.16\%$ and $8.26\%$ in the MedDG dataset when considering the ROUGE metric. Furthermore, LLM-based models consistently outperform other models in both the MedDG and KaMed datasets according to the DISTINCT-2 metric. These findings highlight the strength of LLM-based models in generating responses that closely align with the gold standard responses in terms of recall and content coverage, as indicated by the ROUGE metric. Additionally, they demonstrate the ability to produce diverse and distinct responses, as indicated by the DISTINCT-2 metric. Conversely, other models may prioritize response quality based on precision and n-gram matching, as indicated by their performance in the BLEU metric. In summary, the results underscore the strengths of LLM-based models in generating high-quality responses that capture both the richness and diversity in MDG, making them particularly suitable for tasks requiring comprehensive and diverse outputs.

\subsection{Ablation Study}
To assess the impact of different modules in BP4ER, we conducted ablation studies on two datasets by individually removing modules PR-Bootstrap, AP-Bootstrap, and the explicit reasoning process, as outlined in Table \ref{ablation}.

Firstly, we analyzed the effects of PR-Bootstrap on performance. Comparing the results to BP4ER, we observed decreases in all metrics upon removing PR-Bootstrap. This suggests that instructing the model of its incorrectness by revising the prompt positively influences model performance. Secondly, when removing AP-Bootstrap from BP4ER, we notice a slight increase in the DISTINCT-1/2 metric. We hypothesize that this improvement may be attributed to the fact that AP-Bootstrap can be considered as a form of greedy decoding, which tends to generate repetitive or monotonous sequences rather than diverse content. Finally, upon removing the explicit reasoning process, we observed a decline in all evaluation metrics, with a notably significant drop of $8.3\%$ in the ROUGE metric in the MedDG dataset. This indicates that the introduction of an explicit reasoning process enhances the interpretability of the response generation in the MDG and improves the semantic similarity between the generated response and the ground truth.

Our ablation experiments robustly confirm the effectiveness of each module on model performance. The results indicate that all these modules contribute positively to our approach, underscoring their importance in achieving superior performance in MDG tasks.

\subsection{Human Evaluation}
In addition to quantitative evaluations, we conducted a human evaluation to assess the responses generated by different models in terms of fluency, consistency, and entity correctness. We randomly sampled 100 instances from the test set of KaMed, and the corresponding responses were generated by well-performing models such as DialoGPT, MedPIR, and BP4ER. To ensure the fairness of the assessment, the responses for each sample were shuffled and then presented to volunteers for evaluation. The final statistical results are summarized in Table \ref{human_eval}. Notably, our proposed model BP4ER consistently outperformed other models across all three manual evaluation indicators. Particularly noteworthy is BP4ER's superiority in fluency and coherence, suggesting that our proposed method significantly enhances the quality of responses. This improvement can be attributed to the explicit decomposition of MDG's multi-step reasoning process and the iterative bootstrapping on prompting, both of which contribute to generating more linguistically fluent and contextually coherent responses, improving the overall user experience in medical dialogue scenarios. These findings provide additional evidence of the effectiveness of BP4ER in improving the quality of responses in MDG tasks, demonstrating its superiority over existing models.

\begin{figure}
\begin{center}
\includegraphics[scale=0.42]{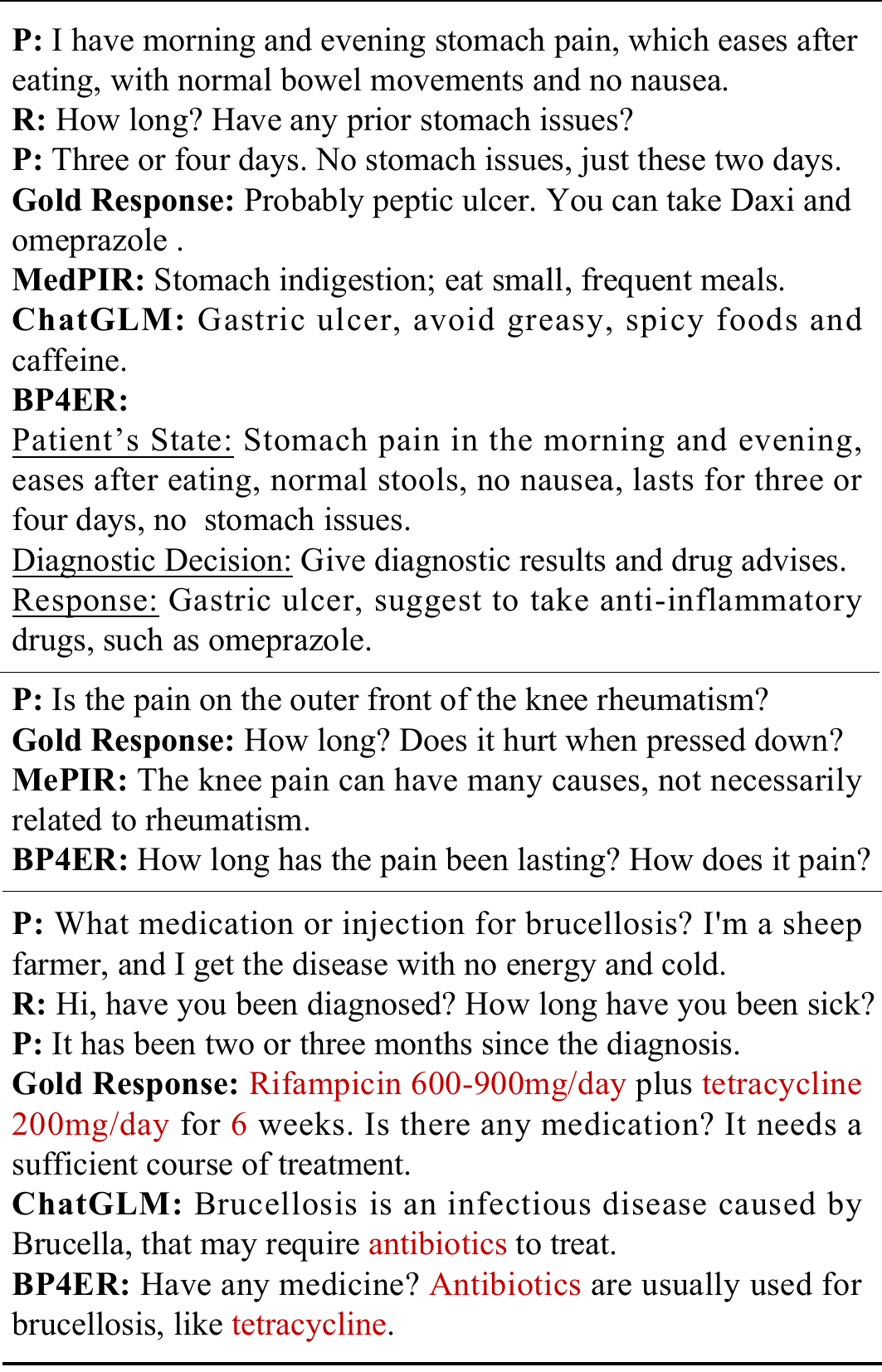} 
\caption{A case study on comparative responses generated from various models, where "P" represents patient descriptions and "R" represents system responses.}
\label{fig_case}
\end{center}
\end{figure}

\subsection{Case Study}
We randomly selected dialogue examples from the KaMed test set and compared the generated responses from several models in Figure \ref{fig_case}.

In case 1, BP4ER follows a multi-step reasoning procedure: firstly, it summarizes the patient's current state and then infers the next diagnostic action for the physician, i.e., inquire about the patient's medical history. Finally, BP4ER generates a response semantically close to the gold response, both emphasizing "asking for the specific area of pain". In contrast, other models primarily focus on providing suggestions. This highlights BP4ER's ability to produce responses that are more aligned with the context of the dialogue, achieved through explicit reasoning in MDG and iterative bootstrap prompting.

In case 2, upon receiving the patient's self-reported information, BP4ER predicts the next diagnostic decision and generates a response to inquire about the patient's drug history. This rationale closely aligns with medical logic. Conversely, other models offer advice without a comprehensive understanding of the patient's medical background, lacking medical rationale.

In case 3, it is apparent that ChatGLM-6B is limited to providing only approximate antibiotic drug recommendations. Conversely, BP4ER exhibits a more advanced ability by not only inquiring about the patient's medication history but also providing specific recommendations for antibiotic drugs suitable for the individual's condition. Despite this advanced ability, BP4ER still falls short when compared to the gold standard response, particularly in accurately determining the appropriate dosage and duration for medication use. This finding underscores the crucial necessity of integrating expert medical knowledge into the model to achieve precision for effective medical decision-making.

\section{Conclusion}
In this paper, we propose BP4ER, a novel medical dialogue generation (MDG) model. BP4ER employs a least-to-most prompting strategy to guide a large language model (LLM) towards explicit reasoning. This strategy involves breaking down MDG into a sequence of interrelated sub-questions, making the process closer to real medical reasoning. Each sub-question is driven by answers obtained from resolving preceding queries. Furthermore, the model incorporates two iterative bootstrapping techniques for prompting, enhancing the LLM's explicit reasoning ability. Through the iterative approach, BP4ER autonomously corrects intermediate errors, leading to more precise and coherent medical responses. These features collectively enhance the transparency and interpretability of the medical reasoning process while improving the overall quality of the generated medical dialogue responses. Both automatic and human evaluations consistently show BP4ER outperforming existing state-of-the-art methods.

\section{Limitations}
Given that the BP4ER relies on large language models and prompts to direct response generation, it necessitates greater computational resources to execute the reasoning chain and bootstrap prompting prior to generating responses. Another crucial limitation lies in the potential for the model to generate incorrect or nonsensical responses during the reasoning process. This risk arises from the inherent reliance on the reasoning ability of LLMs, which possess general knowledge but lack the specialized medical knowledge for accurate medical dialogue generation. As a result, there's a notable gap between the model's understanding of medical concepts. In future work, we hope to explore the introduction of medical knowledge to further enhance the model's explicit reasoning ability in the medical domain.

\section{Acknowledgements}
This work was supported by the National Key Research and Development Program of China under Grant 2021YFE0205700, Beijing Natural Science Foundation JQ23016, the External cooperation key project of Chinese Academy Sciences 173211KYSB20200002, the Science and Technology Development Fund of Macau Project 0123/2022/A3, and 0070/2020/AMJ, the National Natural Science Foundation of China (No.62376270), Open Research Projects of Zhejiang Lab No. 2021KH0AB07 and CCF-Zhipu AI Large Model Project 202219.

\nocite{*}
\section{Bibliographical References}\label{sec:reference}

\bibliographystyle{lrec-coling2024-natbib}
\bibliography{lrec-coling2024-example}

\label{lr:ref}

\end{document}